\pdfoutput=1

\documentclass[11pt]{article}

\usepackage[preprint]{acl}

\usepackage{tikz}
\usetikzlibrary{bayesnet}
\usepackage{times}
\usepackage{blindtext}
\usepackage{latexsym}
\usepackage{amsmath}
\usepackage{makecell}
\usepackage{amssymb}
\usepackage{dsfont}
\usepackage{graphicx}
\usepackage{subcaption}
\usepackage{threeparttable}
\usepackage{algorithm}
\usepackage{algorithmic}
\usepackage{float}
\usepackage{enumitem} 

\usepackage[T1]{fontenc}

\usepackage[utf8]{inputenc}

\usepackage{microtype}

\usepackage{inconsolata}

\usepackage{graphicx}

\usepackage{multirow}
\usepackage[normalem]{ulem}
\useunder{\uline}{\ul}{}

\title{GloCOM:  A Short Text Neural Topic Model via Global Clustering Context}

\author{
 \textbf{Quang Duc Nguyen\textsuperscript{1}},
 \textbf{Tung Nguyen\textsuperscript{1}},
 \textbf{Duc Anh Nguyen\textsuperscript{1}},
\\
 \textbf{Linh Van Ngo\textsuperscript{1}\textsuperscript{$\dagger$}},
 \textbf{Sang Dinh\textsuperscript{1}},
 \textbf{Thien Huu Nguyen\textsuperscript{2}}
\\
 \textsuperscript{1}Hanoi University of Science and Technology, Vietnam \\
 \textsuperscript{2}University of Oregon, USA
 }

\begin{document}
\maketitle

\renewcommand{\thefootnote}{$\dagger$}
\footnotetext[1]{Corresponding author: \href{mailto:email@domain}{linhnv@soict.hust.edu.vn}} 
\renewcommand*{\thefootnote}{\arabic{footnote}}

\begin{abstract}

    Uncovering hidden topics from short texts is challenging for traditional and neural models due to data sparsity, which limits word co-occurrence patterns, and label sparsity, stemming from incomplete reconstruction targets. Although data aggregation offers a potential solution, existing neural topic models often overlook it due to time complexity, poor aggregation quality, and difficulty in inferring topic proportions for individual documents. In this paper, we propose a novel model, \(\textbf{GloCOM}\) (\textbf{Glo}bal \textbf{C}lustering C\textbf{O}ntexts for Topic \textbf{M}odels), which addresses these challenges by constructing aggregated global clustering contexts for short documents, leveraging text embeddings from pre-trained language models. GloCOM can infer both global topic distributions for clustering contexts and local distributions for individual short texts. Additionally, the model incorporates these global contexts to augment the reconstruction loss, effectively handling the label sparsity issue. Extensive experiments on short text datasets show that our approach outperforms other state-of-the-art models in both topic quality and document \(\text{representations}\).\footnote{Our code will be publicly available upon publication.}
    
\end{abstract}

\section{Introduction}\label{section:introduction}

        
    Topic models~\cite{1999plsi, blei2003lda} have proven effective in discovering topics within a corpus and providing a high-level representation of documents. Topic models are applied in various domains, including text mining \cite{2017_effective, 2022podcast}, bioinformatics \cite{2020bioin4}, and recommender systems \cite{2018collaborative} and streaming learning \cite{nguyen2019infinite,van2022graph,nguyen2021boosting}.  However, while they perform well with long texts, these models often struggle with short data~\cite{tuan2020bag,bach2020tps,ha2019eliminating,nguyen2022adaptive}. Datasets containing short documents, such as headlines, comments, or search snippets, offer limited information on word co-occurrence~\cite{qiang2018STTP}, essential for identifying latent topics. This challenge, known as data sparsity, significantly hinders the ability of recent models to generate high-quality topics. Moreover, the brevity of short texts introduces label sparsity~\cite{lin-etal-2024-combating}, where unobserved but relevant words are ignored in the evidence lower bound, causing biased reconstruction loss in Variational Autoencoder (VAE)-based neural topic models~\cite{kingma2013vae}.

    Document aggregation has effectively addressed short text topic modeling challenges~\cite{Hong_2010, Quan_aggregate_2015}. However, modern neural network-based topic models~\cite{wu-etal-2020-short, wu-etal-2022-mitigating, lin-etal-2024-combating} have not paid much attention to this approach due to the limitations demonstrated in traditional research. Particularly, aggregation approaches that rely on auxiliary information~\cite{Hong_2010, Mehrotra_tweet_2013} are often restricted to specific data types, while self-aggregation methods face challenges such as high time complexity or overfitting as data volume increases~\cite{Quan_aggregate_2015, Zuo2016TopicMO}. Moreover, some methods are unable to infer topic distributions for individual documents~\cite{Weng_twitter_2010, Tang_shorttm_2013}. 
    
    The kNNTM~\cite{lin-etal-2024-combating} model is the first short text neural topic model to address label sparsity using kNN-based document aggregation. By enhancing the reconstruction target with semantically related documents, the model can leverage word co-occurrence patterns and the relationships between documents in the dataset. Although this approach has proven effective, kNNTM still faces significant time costs due to optimal transport measures between every pair of documents in the corpus. Another natural and cost-effective method for aggregation is through clustering, but data sparsity is also an unavoidable issue for clustering algorithms based on traditional text representation with term frequency~\cite{Quan_aggregate_2015, Jin_cluster_2011}.

      \begin{table}
        \centering
        \setlength{\tabcolsep}{0.8mm}
         \renewcommand{\arraystretch}{1.1}
        \resizebox{\linewidth}{!}{
        \begin{tabular}{l}
        \Xhline{1.15pt}
        \textbf{Cluster 1:}  \\
        \begin{tabular}[c]{@{}l@{}}\#1: nokia lumia launch\\
        \#2: moto order start shipping december support \\
        \#3: microsoft officially launched xbox console gamers  \end{tabular} \\ \hline
        \textbf{Cluster 2:} \\
        \begin{tabular}[c]{@{}l@{}}\#4: black friday sale cyber monday\\
        \#5: local shop start sale thanksgiving day\\
        \#6: shopper black friday\end{tabular}                      \\ \Xhline{1.15pt}
        \end{tabular}
        }
        \caption{Examples of short texts with their corresponding clustering contexts from the GoogleNews~\cite{yin2016model} dataset using a PLM embeddings model, \texttt{all-MiniLM-L6-v2}~\cite{reimers-gurevych-2019-sbert}. The global topic distributions might have high probabilities for topics like \textit{Technology} and \textit{Sales} for Clusters 1 and 2, respectively.}
        \label{tab:intro-cluster-example}
    \end{table}
    
    To address these issues, we propose a novel topic model called GloCOM, which constructs a Neural Topic Model using Global Clustering Context. Our model first clusters the document set and then creates global contexts (or global documents) by merging short documents (or local documents) within each cluster. The topic model then incorporates global and local texts, yielding global and local topic distributions. Each local topic distribution is derived from the global distribution of its corresponding context, supplemented by a learnable noise parameter. This approach enables the model to harness the benefits of document aggregation to enhance word co-occurrences and handle the data sparsity issue while inferring representations for individual documents. Furthermore, advancements in large language models~\cite{devlin2019bert, gpt3} have improved short text processing effectiveness for downstream tasks. Consequently, we employ pre-trained language model (PLM) embeddings~\cite{reimers-gurevych-2019-sbert, behnamghader2024llm2veclargelanguagemodels} to represent texts during the clustering process. As shown in Table~\ref{tab:intro-cluster-example}, the effectiveness of PLM-based global clustering can thus enhance the semantic quality of the topics.


    Besides, we augment the model's reconstruction targets by combining the original short text with the globally aggregated documents mentioned above. This approach allows us to globally capture absent yet relevant words related to the original input, such as ``shop'' and ``shopper'' in documents \#5 and \#6, which belong to the same cluster, as shown in Table~\ref{tab:intro-cluster-example}. This strategy effectively addresses the issue of label sparsity, where the probabilities of related, unobserved words are inappropriately reduced, resulting in biased training signals and producing high-quality topics and document-topic distributions. Moreover, compared to the considerable time demands of kNNTM~\cite{lin-etal-2024-combating}, our method, using cluster context augmentation, is far more time- and resource-efficient. We summarize our contributions as follows:
    \vspace*{-4pt}
    \begin{itemize}[noitemsep,left=0.5pt]
    \item We propose a novel Neural Topic Model for short texts, named GloCOM, which addresses data sparsity by leveraging pre-trained language model embeddings to construct global clustering contexts for documents.
    \item We introduce a novel inference mechanism that captures both global and local document-topic distributions, which enhances the representation of individual short documents.
    \item We enhance the VAE's reconstruction loss by integrating short texts with global clustering contexts, allowing the model to capture unobserved yet relevant words and improve topic quality. 
    \item We conduct extensive experiments on benchmark datasets, demonstrating that GloCOM outperforms existing models in terms of topic coherence and document-topic distribution.
\end{itemize}
    
\section{Preliminaries} \label{sec:preliminaries}

\subsection{Notations} \label{subsec:notations}

    Denote $\mathbf{X}=\{x^d\}_{d=1}^{D}$ as a collection of Bag-of-Words (BoW) representations of $D$ documents with the vocabulary of $V$ words. Topic models aim to discover $K$ hidden topics in this corpus. The pre-trained language model embedding of document~$d$ is $x^d_{PLM}$. The clustering algorithm applied to $x^d_{PLM}$ produces $G$ clusters. We have $\beta \in \mathbb{R}^{V\times K}$ = $(\beta_1,\dots,\beta_K)$, where each $\beta_k \in \mathbb{R}^{V\times 1}$, as the topic-word distributions of $K$ desired topics. 
    
    With $L$ as the word embedding dimension, we set $ \textbf{w}_v \in \mathbb{R}^{L}, v \in \{1,2,\dots,V\}$ and ${\textbf{t}_k \in \mathbb{R}^{L}}, k \in \{1,2,\dots,K\}$ to be the word embeddings of word $v$ and topic embeddings of topic $k$, respectively. Each document $x^d$ has the topic proportion  $\theta_d \in \mathbb{R}^{K}$ indicating what topic it includes. $\mathds{1}_N$ denotes a vector of length $N$, where each entry is set to 1.

\subsection{VAE-based Topic Model} \label{subsec:VAE-based Topic Model}
    
    Similar to many recent neural topic models~\cite{dieng2020etm, wu2023effective}, our approach is built on a VAE framework, which consists of two primary components: (i) an inference encoder that produces document-topic distributions; and (ii) a generative decoder that reconstructs the original text using the encoder’s output and the topic-word proportions. For the encoder, the Bag-of-Words (BoW) representation of a document $x^d$ is processed through neural networks to obtain the parameters of a normal distribution, where the mean $\mu=h_\mu(x^d)$ and the diagonal covariance matrix $\Sigma=\mathrm{diag}(h_\Sigma(x^d))$ are computed. The reparameterization trick~\cite{kingma2013vae} is then employed to sample a latent variable $\alpha$ from the posterior distribution $q(\alpha \vert x)=\mathcal{N}(\alpha \vert \mu, \Sigma)$, while the prior distribution of $\alpha$ is $p(\alpha)=\mathcal{N}(\alpha \vert \mu_0, \Sigma_0)$. Afterwards, the softmax function is applied to $\alpha$, producing the topic proportion $\theta=\mathrm{softmax}(\alpha)$. 
        
    Regarding the second component, VAE-based neural topic models aim to construct an effective representation for the topic-word distributions $\beta \in \mathbb{R}^{V\times K}$. There are several approaches to modeling $\beta$, such as directly inferring it through an optimization process~\cite{srivastava2017prodlda} or decomposing $\beta$ into the product of word embeddings~$\mathcal{W}$ and topic embeddings~$\mathcal{T}$. Alternatively,~\cite{wu2023effective} propose another form of $\beta$ that effectively addresses the issue of topic collapse as follows:
    \begin{equation} \label{eq:beta_decomposition}
            \beta_{ij} = \frac{\exp \left(-\Vert \mathbf{w}_i - \mathbf{t}_j \Vert^2 / \tau \right)}{\sum_{j'=1}^K \exp \left(-\Vert \mathbf{w}_i - \mathbf{t}_{j'} \Vert^2 / \tau \right)},
    \end{equation}
    where $\tau$ is a temperature hyperparameter. The word embeddings $\mathcal{T}$ are typically initialized using pre-trained embeddings such as GloVe~\cite{pennington-etal-2014-glove}.


     \begin{figure}
        \centering
        \includegraphics[width=0.48\textwidth]{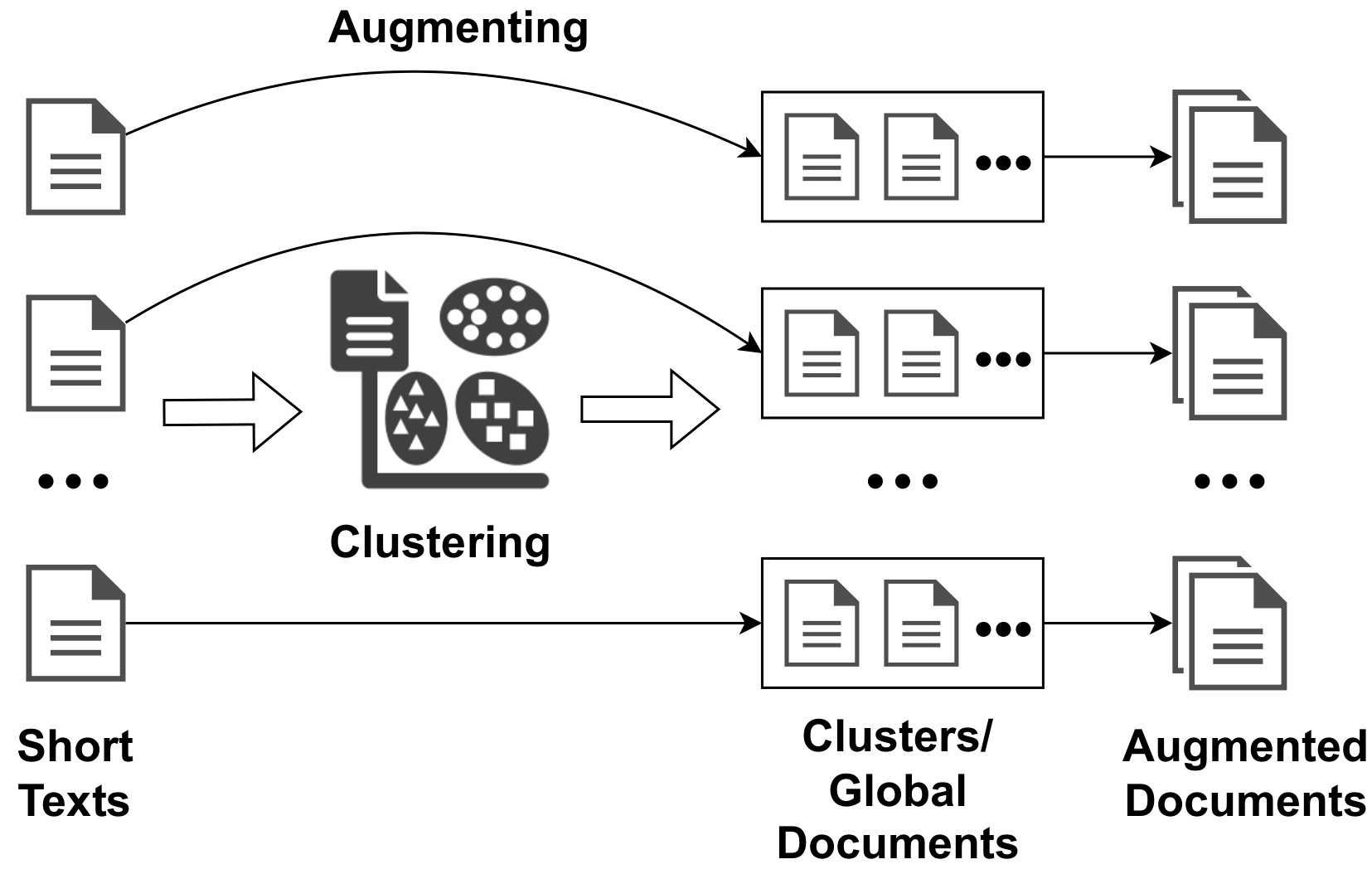}
        \caption{Our short text aggregation illustration. We cluster short texts using PLM embeddings and form global documents by concatenating texts from each cluster. Each text is then augmented with its corresponding global document, creating an augmented document used in the reconstruction loss.}
        \label{fig:text-aggregation}
    \end{figure}
    
    VAE-based models aim to reconstruct the BoW representations of documents using the topic-word distribution matrix $\beta$ and document-topic proportion $\theta_d$ as $\hat{x}^d \sim \mathrm{ Multinomial}(\mathrm{softmax}(\beta\theta_d))$. The topic modeling loss consists of a reconstruction term and a regularization term, as follows:
    \begin{equation*} \label{eq:tm}
        \begin{split}
        \mathcal{L}_{\mathrm{TM}} = \frac{1}{D} \sum_{d=1}^{D} \Big[& - (x^d)^{\top} \log (\mathrm{softmax}(\beta \theta_d)) \\ &+ \mathrm{KL}(q(\alpha \vert x^d) \| p(\alpha)) \Big].
        \end{split}
    \end{equation*}

    Furthermore, we leverage the Embedding Clustering Regularization loss $\mathcal{L}_{\mathrm{ECR}}$~\cite{wu2023effective} to handle the topic repetition problem, which is prevalent issues in short text topic models, as highlighted in previous studies~\cite{wu-etal-2020-short, wu-etal-2022-mitigating}. We provide details about Embedding Clustering Regularization in Appendix~\ref{appendix:ecr}.

\section{Methodology}\label{sec:method}

    We propose a novel topic model framework for short texts, which introduces an effective cluster-based text aggregation method and an innovative inference for both the global enhanced context and each individual document. Furthermore, we reuse the global context to augment the VAE's reconstruction labels, addressing the label sparsity issue.
    
\subsection{Short Text Aggregation via PLM-based Clustering} \label{subsec:short-text-aggre}

    Modern neural topic models for short texts do not focus heavily on the text aggregation approach because traditional term frequency representations fail to effectively capture word co-occurrence within a group of short texts~\cite{Quan_aggregate_2015}. To overcome this limitation, we utilize pre-trained language model embeddings~\cite{reimers-gurevych-2019-sbert, behnamghader2024llm2veclargelanguagemodels}, which excel at capturing linguistic patterns and contextual nuances, to represent texts for clustering. These enriched embeddings facilitate document clustering by grouping texts with similar semantic meanings.
        
    Our short text aggregation process is described in Figure~\ref{fig:text-aggregation}. We concatenate short texts (local documents) within the same cluster, forming what we refer to as a global document $x^g$, with $g$ being the cluster containing document $x^d$. To address the label sparsity problem, we construct the augmented documents as $\Tilde{x}^d = x^d + \eta x^g$, where $\eta$ is the augmentation coefficient.
    
    While standard topic models are typically applied directly to the global documents to extract topics, our study introduces a novel approach: we estimate both global and local topic distributions for the corresponding global and local documents, with the reconstruction loss built using the augmented documents.

\subsection{Global Clustering Context Topic Model} \label{subsec:global-clustering-context}

\begin{figure}
        \centering
        \resizebox{0.48\textwidth}{!}{ 
          \begin{tikzpicture}
            \node[obs] (alpha) {$\mathcal{L}\mathcal{N}(.)$} ; %
            \node[latent, right= 0.8 of alpha] (theta^g) {$\theta^g$} ; %
            \node[latent, right= 0.8 of theta^g] (theta^g_d) {$\theta^g_d$} ; %
            \node[latent, above= 0.5 of theta^g_d] (rho_d) {$\rho_d$} ; %
            \node[obs, above= 1.0 of rho_d] (noise_prior) {$\epsilon$} ; %
            \node[latent, right= 0.8 of theta^g_d] (z) {$z$} ; %
            \node[obs, right= 0.5 of z] (w) {$w$} ; %
            \node[latent, right= 1.4 of w] (beta) {$\beta$} ; %
            \node[latent, below= 0.5 of beta] (word) {$\mathcal{W}$};%
            \node[latent, above= 0.5 of beta] (topic) {$\mathcal{T}$};%
            \plate[inner xsep=0.35cm, inner ysep=0.2cm, xshift=0cm, yshift=0.15cm] {plate1} { (z) (w)} {$N_d$}; %
            \plate[inner xsep=0.5cm, inner ysep=0.2cm, xshift=0cm, yshift=0.12cm] {plate2} {(rho_d) (theta^g_d) (plate1)} {$D_g$}; %
            \plate[inner xsep=0.3cm, inner ysep=0.2cm, xshift=0cm, yshift=0.15cm] {plate3} {(theta^g) (plate2)} {$G$}; %
            \edge {alpha} {theta^g} ; %
            \edge {theta^g} {theta^g_d} ; %
            \edge {theta^g_d} {z} ; %
            \edge {z,beta} {w} ; %
            \edge {word} {beta};
            \edge {topic} {beta};
            \edge {rho_d} {theta^g_d};
            \edge {noise_prior} {rho_d}
          \end{tikzpicture}
        }
        \caption{The probabilistic graphical model illustrating the generative process of documents in \text{GloCOM}.}
        \label{fig:GloCOM-generative}
    \end{figure}
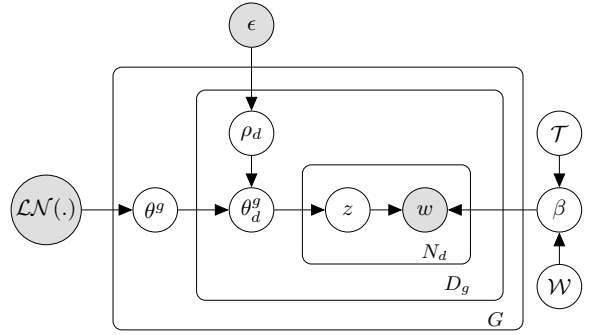

    Now, we provide formal descriptions of GloCOM. Let $\theta^g \in \mathbb{R}^{K}$ denote the topic distribution of the global document $x^g$. We introduce a latent adaptive variable, $\rho_d \in \mathbb{R}^{K}$, which controls the extent to which $\theta^g$ influences the topic proportions of each individual document within the cluster. Using both $\theta^g$ and $\rho_d$, we construct the topic distribution for each short text, denoted as $\theta^g_d \in \mathbb{R}^{K}$. Under the GloCOM, the generative process for the documents (as illustrated in Figure \ref{fig:GloCOM-generative}) is as follows:

    \begin{enumerate}
    \item Calculate the distribution over words $\beta$ as described in Eq.~\ref{eq:beta_decomposition}.
    \item For each cluster $g$ : Generate $\theta^g \sim \mathcal{L}\mathcal{N}(0, I)$, with $\mathcal{L}\mathcal{N}$ denotes logistic-normal
    distribution.
    \item For each document $d$ in cluster $g$: 
        \begin{enumerate}
            \item Draw adaptive variable $\rho_d \sim \mathcal{N}(1, \epsilon I)$, where $\epsilon$ is a hyperparameter.
            \item Generate topic distribution  
            {\abovedisplayskip=0pt \belowdisplayskip=0pt
            \begin{align} \label{eq:generate-topic-distribution}
            \theta^g_d = \mathrm{softmax} (\theta^g \odot \rho_d).
            \end{align}
            }
            \item For each $n^{th}$ word in document $d$:
                \begin{enumerate}
                    \item Draw a topic index: 
                    
                    $z_{dn} \sim \mbox{Multinomial}(\theta^g_d)$.
    				\item Draw the word: 
        
                    $w_{dn} \sim \mbox{Multinomial}(\beta_{z_{dn}})$.
                \end{enumerate}
        \end{enumerate}
    
    \end{enumerate}
    
    In Step 1, we define the topic-word distributions $\beta$ using Eq.~\ref{eq:beta_decomposition}. This formulation captures the clustering relationships between topics and word embeddings~\cite{wu2023effective}, effectively addressing the issue of topic repetition, which is particularly problematic for short text topic models due to limited word co-occurrences~\cite{wu-etal-2020-short, wu-etal-2022-mitigating}.
    
    Steps 2, 3a, and 3b introduce key innovations for short text topic modeling in our approach. We treat the global context within a cluster as a single long document, generating its topic distribution. The global context serves as a semantic representation for the entire cluster, which consists of multiple sub-documents, and enhances word co-occurrence in short text corpora. By leveraging the topic distribution from this global document, we generate topic proportions for each sub-document through an adaptive variable, $\rho_d$, specific to each short text.


    Step 3c follows most topic modeling approaches \cite{blei2003lda, srivastava2017prodlda, dieng2020etm}. In this step, a topic assignment is first sampled for each word, and then the topic-word distribution is used to generate the words for the document. What sets this process apart is that the topic assignments are drawn from $\theta^g_d$, a newly designed topic proportion for short texts that captures both the cluster's global semantic context and the individual document's specific information.

\subsection{Inference and Estimation}\label{subsec:inference}

    \begin{figure*}
        \centering
        \includegraphics[width=0.89\textwidth]{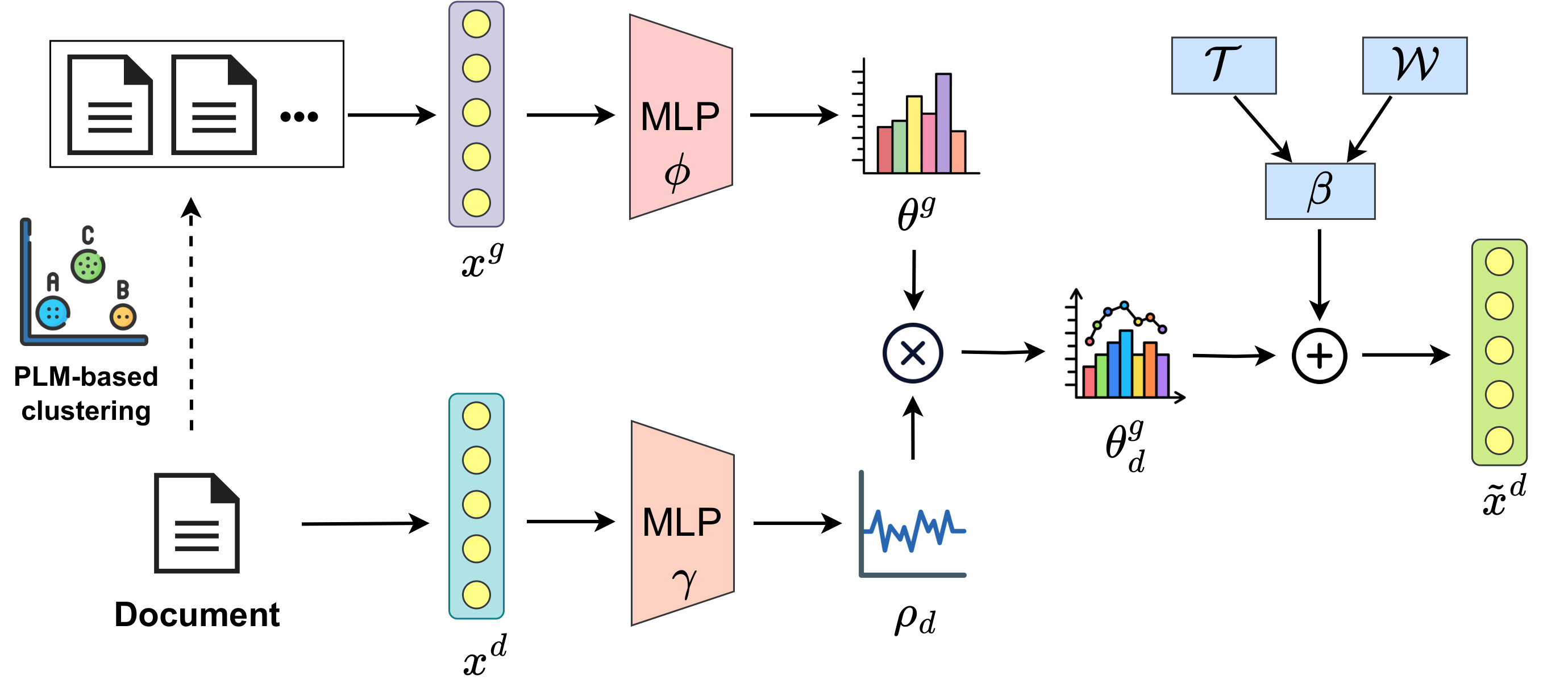}
        \caption{The overall architecture of GloCOM. Our methods generate global and augmented documents from clustering based on pre-trained language model embeddings. GloCOM proposes a novel approach to estimate both global and local doc-topic distributions and incorporates the augmented documents into the reconstruction loss.}
        \label{fig:model}
    \end{figure*}

    The marginal likelihood over dataset $\mathbf{X}$ is composed of a sum over the marginal likelihoods of individual documents $x^d$. Here, we also consider the group of texts, the expansion of the marginal likelihood can be expressed as:
    {\abovedisplayskip=0.25pt \belowdisplayskip=0.25pt
    \begin{equation*}
        \label{eq:expansion}
        \begin{split}
            \log p(\mathbf{X}| w, t, \epsilon)\!&=\!\!\!\sum_{d}^{D} \!\sum_{g}^{G}\!\mathbb{I}[x^d \in g]\!\log p(x^d | w, t, \epsilon), 
        \end{split}
    \end{equation*}
    }
    where $\mathbb{I}[x^d \in g]$ equals $1$ if $x^d \in g$ and equals to $0$ otherwise. The challenge is that the marginal likelihood of each document is intractable to compute, due to the difficult integral of $\log p(x^d | w, t, \epsilon)$ over the global topic distribution $\theta^g$ and the adaptive variable $\rho_d$: $\int \int p(\theta^g, \rho_d | x^d, \epsilon) p(x^d | \theta^g, \rho_d, w, t) \, \mathrm{d} \theta^g \mathrm{d} \rho_d$.
    
    To solve this problem, we introduce the variational distribution $q_{\phi, \gamma}(\theta^g, \rho_d | x^g, x^d)$, which is the approximation to the intractable true posterior $p(\theta^g, \rho_d | x^d, \epsilon)$. For greater simplification, we use the fully factorized assumption in the variational distribution: $q_{\phi, \gamma}(\theta^g, \rho_d | x^g, x^d) = q_\phi(\theta^g | x^g) q_\gamma(\rho_d | x^d)$. We then employ amortized inference for both $\theta^g$ and $\rho_d$, where their variational distribution depends on the data and the shared variational parameters and the shared variational parameters $\phi$ and $\gamma$, respectively. Specifically, both $q_\phi(\theta^g | x^g)$ and $q_\gamma(\rho_d | x^d)$ are modeled as Gaussian distributions, with their means and variances generated from inference neural networks parameterized by $\phi$ and $\gamma$. However, while the network for $\rho_d$ processes $x^d$ itself, the input to the network for $\theta^g$ is the global document $x^g$, which includes $x^d$.

    The marginal likelihood of each data point can be written as:
    \begin{equation*}
        \label{eq:elbo_generate}
        \begin{split}
            &\log p(x^d | w, t, \epsilon) = \mathcal{L}^d (\phi,\gamma,w,t) \\
            &+D_{KL} \left(q_{\phi, \gamma}(\theta^g, \rho_d | x^g, x^d) || p(\theta^g, \rho_d | x^d, \epsilon)\right) \\ 
        \end{split}
    \end{equation*}
    Since the $D_{KL}$ is non-negative, the term $\mathcal{L}^d (\phi,\gamma,w,t)$ is called the variational lower bound on the marginal likelihood of document $d$, and can be expanded as:
    \begin{align*} 
        &\log p(x^d | w, t, \epsilon) \geq \mathcal{L}^d (\phi,\gamma,w,t) \notag \\
        &= \mathbb{E}_{q_{\phi, \gamma}(\theta^g, \rho_d | x^g, x^d)} \left[ \log p(x^d, \theta^g, \rho_d | w, t, \epsilon) \right] \notag \\
        &\quad - \mathbb{E}_{q_{\phi, \gamma}(\theta^g, \rho_d | x^g, x^d)} \left[ \log q_{\phi, \gamma}(\theta^g, \rho_d | x^g, x^d) \right] \notag \\
        &= \mathbb{E}_{q_\phi(\theta^g | x^g) q_\gamma(\rho_d | x^d)} \left[ \log p(x^d | \theta^g, \rho_d, w, t) \right] \notag \\
        &\quad - D_{KL} \left(q_\phi(\theta^g | x^g) || p(\theta^g)\right) \notag \\
        &\quad - D_{KL} (q_\gamma(\rho_d | x^d) || p(\rho_d | \epsilon))
    \end{align*}
    Optimizing the lower bound can lead to better approximations of the marginal likelihood and ensure that the approximate posterior $q_{\phi, \gamma}(\theta^g, \rho_d | x^d)$ closely resembles the true posterior distribution. The first component of the lower bound represents the reconstruction loss, which aims to recreate the input data. The remaining components serve as regularizers, promoting the alignment of global and local topic distributions with their prior distributions. To ensure that the lower bound remains tractable and differentiable, we employ Monte Carlo approximations and the reparameterization trick~\cite{kingma2013vae}. We then use the augmented reconstruction label $\Tilde{x}^d$, as discussed in Section~\ref{subsec:short-text-aggre}. Finally, the lower bound for document $d$ can be expressed as:
    \begin{align}
        \mathcal{L}^d (\phi,\gamma,w,t) 
        &= - (\Tilde{x}^d)^{\top} \log \big(\mathrm{softmax}(\beta \theta^g_d)\big) \notag \\
        &- D_{KL} \left(q_\phi(\theta^g | x^g) || p(\theta^g)\right) \notag \\
        &- D_{KL} (q_\gamma(\rho_d | x^d) || p(\rho_d | \epsilon)) 
        \label{eq:final_los}
    \end{align}
    
    We consider the marginal likelihood lower bound of the full dataset as $\mathcal{L}_{TM}$, which is expressed as follows:
    \begin{equation}
    \label{eq:overall_elbo}
    \begin{split}
        \mathcal{L}_{\mathrm{TM}} = \sum_d^D \sum_g^G \mathbb{I}[x^d \in g] \mathcal{L}^d (\phi,\gamma,w,t)
    \end{split}
    \end{equation}

    To mitigate the topic collapse problem, we apply Embedding Clustering Regularization, as detailed in Section~\ref{sec:preliminaries}. In summary, the overall objective function of GloCOM is described below:
    \begin{equation}
    \label{eq:overall}
    \begin{split}
        \mathcal{L}_{\mathrm{GloCOM}} = \mathcal{L}_{\mathrm{TM}} + \lambda_{\mathrm{ECR}} \mathcal{L}_{\mathrm{ECR}}
    \end{split}
    \end{equation}
    where $\lambda_{\mathrm{ECR}}$ is a weight hyperparameter. The overall architecture of GloCOM shown in Figure~\ref{fig:model}. We also provide the training algorithm in Appendix~\ref{appendix:algorithm}.

    \begin{table*}
        \centering
        \setlength{\tabcolsep}{0.8mm}
        \renewcommand{\arraystretch}{1.2}
        \resizebox{\linewidth}{!}{
        \begin{tabular}{lcccccccccccccccccccc}
        \hline
        \multirow{2}{*}{{\textbf{\begin{tabular}[l]{@{}l@{}}Model\\ $K = 50$\end{tabular}}}} & & \multicolumn{4}{c}{\textbf{GoogleNews}} & & \multicolumn{4}{c}{\textbf{SearchSnippets}} & & \multicolumn{4}{c}{\textbf{StackOverflow}} & & \multicolumn{4}{c}{\textbf{Biomedical}} \\ \cline{3-6} \cline{8-11} \cline{13-16} \cline{18-21}
                                        & & TC          & TD          & Purity      & NMI         & & TC          & TD          & Purity      & NMI         & & TC          & TD          & Purity      & NMI         & & TC          & TD          & Purity      & NMI         \\ \hline
        ProdLDA                         & & 0.437  & 0.991  & 0.201  & 0.384  & & 0.406  & 0.546  & 0.731  & 0.435  & & 0.388  & 0.588  & 0.117  & 0.151  & & 0.469  & 0.520  & 0.136  & 0.177  \\
        ETM                             & & 0.402  & 0.916  & 0.366  & 0.560  & & 0.397  & 0.594  & 0.688  & 0.389  & & 0.367  & 0.766  & 0.418  & 0.280  & & 0.450  & 0.723  & 0.406  & 0.273  \\
        ECRTM                           & & 0.441  & 0.987  & 0.396  & 0.615  & & 0.450  & \textbf{0.998} & 0.711  & 0.419  & & 0.381  & 0.941  & 0.197  & 0.192  & & 0.468  & 0.987  & 0.414  & 0.315  \\
        FASTopic                        & & 0.446  & 0.440  & 0.351  & 0.659  & & 0.395  & 0.710  & 0.792  & 0.481  & & 0.317  & 0.222  & 0.408  & 0.486  & & 0.418  & 0.482  & 0.456  & 0.369  \\
        NQTM                            & & 0.408  & 0.959  & 0.536  & 0.716  & & 0.436  & 0.922  & 0.435  & 0.150  & & 0.382  & 0.933  & 0.392  & 0.238  & & 0.471  & 0.915  & 0.191  & 0.109  \\
        TSCTM                           & & 0.437  & 0.988  & 0.552  & 0.761  & & 0.424  & 0.993  & 0.724  & 0.386  & & 0.378  & 0.911  & 0.572  & 0.418  & & 0.484  & 0.972  & 0.480  & 0.341  \\
        KNNTM                           & & 0.435  & 0.986  & 0.579  & 0.795  & & 0.425  & 0.995  & 0.768  & 0.429  & & 0.380  & 0.922  & 0.636  & 0.490  & & \textbf{0.490} & 0.972  & 0.526  & 0.380  \\ \hline
        \textbf{GloCOM}                 & & \textbf{0.475} & \textbf{0.999} & \textbf{0.586} & \textbf{0.817} & & \textbf{0.453} & 0.956  & \textbf{0.806} & \textbf{0.502} & & \textbf{0.390} & \textbf{0.962} & \textbf{0.653} & \textbf{0.588} & & \textbf{0.490} & \textbf{0.998} & \textbf{0.546} & \textbf{0.437} \\ \hline
          \multirow{2}{*}{{\textbf{\begin{tabular}[l]{@{}l@{}}Model\\ $K = 100$\end{tabular}}}} & & \multicolumn{4}{c}{\textbf{GoogleNews}} & & \multicolumn{4}{c}{\textbf{SearchSnippets}} & & \multicolumn{4}{c}{\textbf{StackOverflow}} & & \multicolumn{4}{c}{\textbf{Biomedical}} \\ \cline{3-6} \cline{8-11} \cline{13-16} \cline{18-21}
                                        & & TC          & TD          & Purity               & NMI         & & TC          & TD          & Purity      & NMI         & & TC                   & TD                   & Purity      & NMI         & & TC                   & TD          & Purity      & NMI         \\ \hline
        ProdLDA                         & & 0.435  & 0.611  & 0.611           & 0.600  & & 0.424  & 0.679  & 0.766  & 0.415  & & 0.382  & 0.466           & 0.098  & 0.090  & & 0.463          & 0.465  & 0.079  & 0.050  \\
        ETM                             & & 0.398  & 0.677  & 0.554           & 0.713  & & 0.389  & 0.448  & 0.692  & 0.365  & & 0.369          & 0.444           & 0.475  & 0.331  & & 0.452          & 0.476  & 0.404  & 0.268  \\
        ECRTM                          & & 0.418  & \textbf{0.991} & 0.342 & 0.491  & & 0.432  & \textbf{0.966} & 0.789  & 0.443  & & 0.375          & \textbf{0.993} & 0.172  & 0.179  & & 0.444          & 0.974  & 0.124  & 0.113  \\
        FASTopic                       & & 0.438  & 0.369  & 0.458           & 0.722  & & 0.386  & 0.634  & 0.807  & 0.458  & & 0.309          & 0.186           & 0.495  & 0.514  & & 0.440          & 0.457  & 0.495  & 0.375  \\
        NQTM                           & & 0.397  & 0.898  & 0.706           & 0.788  & & 0.438  & 0.638  & 0.334  & 0.077  & & 0.379          & 0.818           & 0.417  & 0.255  & & 0.460          & 0.572  & 0.142  & 0.056  \\
        TSCTM                          & & 0.448  & 0.941  & 0.754           & 0.835  & & 0.430  & 0.894  & 0.757  & 0.384  & & 0.380          & 0.620           & 0.563  & 0.386  & & \textbf{0.485} & 0.806  & 0.487  & 0.330  \\
        KNNTM                          & & 0.441  & 0.959  & \textbf{0.797}  & 0.870  & & 0.421  & 0.948  & 0.800  & 0.421  & & 0.381          & 0.663           & 0.611  & 0.436  & & 0.483         & 0.848  & 0.530  & 0.362  \\ \hline
        \textbf{GloCOM}                 & & \textbf{0.450} & 0.944  & 0.761  & \textbf{0.900} & & \textbf{0.443} & 0.920  & \textbf{0.822} & \textbf{0.501} & & \textbf{0.382} & 0.804  & \textbf{0.658} & \textbf{0.585} & & 0.462 & \textbf{0.997}  & \textbf{0.536} & \textbf{0.422} \\ \hline
        \end{tabular}
        }
        \caption{Topic quality, measured using TC and TD, and document-topic distribution quality, assessed with NMI and Purity with $K = 50$ and $K=100$. The \textbf{bold} values indicate the best performance. We report the standard deviation  in Appendix~\ref{appendix:additional-results}.}
        \label{tab:main}
    \end{table*}
    
\section{Experiments} \label{sec:exp}

\subsection{Experimental Setups} \label{subsec:exp-setup}

    \paragraph{Datasets.} We use four real-world short text datasets for our experiment: \textbf{(i)} \textbf{GoogleNews}, which covers 152 main news topics from over 11,000 Google articles; \textbf{(ii)} \textbf{SearchSnippets} consisting of over 12,000 web search results divided into 8 different domains; \textbf{(iii)} \textbf{StackOverflow}, which contains 20,000 question titles from 20 different tags; \textbf{(iv)} \textbf{Biomedical} comprising nearly 20,000 medical articles spread across 20 groups. Please refer to Appendix~\ref{appendix:data-statistics} for the dataset statistics and preprocessing details.

    \footnotetext[2]{We could not find the public codebase, so we implemented it ourselves. Please see more details in Appendix~\ref{appendix:kNNTM-implementation}.}
   \paragraph{Baselines.}We compare our methods with the following baselines in two paradigms. For conventional topic models, we consider (i) ProdLDA~\cite{srivastava2017prodlda}, a pioneering VAE-based topic model; (ii) ETM~\cite{dieng2020etm}, which incorporates word embeddings; (iii) ECRTM~\cite{wu2023effective}, based on ETM with regularization between word and topic embeddings; (iv) FASTopic~\cite{wu2024fastopicfastadaptivestable}, a state-of-the-art model for identifying topics via word, topic, and document embeddings. For short text topic models, we include (vi) NQTM~\cite{wu-etal-2020-short}, a neural topic model dedicated to short text problems with vector quantization for topic distributions; (vii) TSCTM~\cite{wu-etal-2022-mitigating}, adapted from NQTM with an additional contrastive loss on topic distributions; (viii) kNNTM~\cite{lin-etal-2024-combating}, a recent state-of-the-art short text neural topic model that augments a document with its neighbors via the kNN algorithm. Except for kNNTM\footnotemark, we use the implementation of the other models provided by TopMost~\cite{wu2023topmost} and fine-tune these baselines on various datasets and topic numbers.

    \paragraph{Evaluation Metrics.} We follow mainstream studies for short text topic modeling~\cite{wu-etal-2020-short, wu-etal-2022-mitigating, lin-etal-2024-combating} and evaluate the topic quality and document-topic distribution quality. For topic quality, we consider two metrics: \textbf{Topic Coherence (TC)} and \textbf{Topic Diversity (TD)}. Topic Coherence assesses how coherent the identified topics are by examining the co-occurrences of the top words with an external corpus. We adopt a widely-used coherence metric $C_V$, which has shown superior performance compared to other coherence metrics~\cite{2015topiccoherence}. Specifically, we compute $C_V$ with Wikipedia texts as a reference corpus using Palmetto library~\cite{2015topiccoherence}. Furthermore, Topic Diversity measures how distinct the topics are by calculating the proportion of unique top words across the topics. We take the top 15 words for each discovered topic for TC and TD. For the quality of the document-topic distribution, we assess document clustering using \textbf{Purity} and \textbf{NMI}, consistent with previous studies~\cite{zhao2020nstm, wu-etal-2022-mitigating, lin-etal-2024-combating}.

    \paragraph{Implementation Details.} We use the same hyperparameter settings as those in previous state-of-the-art text models~\cite{bianchi-etal-2021-ctm, lin-etal-2024-combating}, including epoch numbers, batch size, learning rate, number of hidden layers, etc. We set \texttt{all-MiniLM-L6-v2}\cite{reimers-gurevych-2019-sbert} as the default pre-trained language model for clustering. Regarding the hyperparameters unique to our method, such as $\epsilon, \eta, \lambda_{\mathrm{ECR}}$, we perform a grid search to identify the optimal values. The remaining detailed settings are described in Appendix~\ref{appendix:model-implementation}.

\subsection{Topic Quality and Document-Topic Distribution Quality Evaluations} \label{subsec:top-quality}

     \begin{table}
        \centering
        \setlength{\tabcolsep}{1.1mm}
        \renewcommand{\arraystretch}{1.25}
        \resizebox{\linewidth}{!}{
        \begin{tabular}{lclccccc}
        \hline
        \multicolumn{3}{c}{\textbf{Method}} & & TC & TD & Purity & NMI \\ \hline
        \multirow{4}{*}{\textbf{\begin{tabular}[c]{@{}l@{}}Google\\ News\end{tabular}}} & & ECRTM & & 0.441 & 0.987 & 0.396 & 0.615 \\
                 & & GloCOM+NoC & & 0.465 & 0.989 & 0.548 & 0.768 \\
                 & & GloCOM+NoA & & 0.436 & \textbf{1.000} & 0.548 & 0.770 \\ \cline{3-8} 
                 & & \textbf{GloCOM} & & \textbf{0.475} & 0.999 & \textbf{0.586} & \textbf{0.817} \\ \hline
        \multirow{4}{*}{\textbf{\begin{tabular}[c]{@{}l@{}}Search\\ Snippets\end{tabular}}} & & ECRTM & & 0.450 & 0.998 & 0.711 & 0.419 \\
                 & & GloCOM+NoC & & 0.449 & 0.855 & 0.805 & 0.484 \\
                 & & GloCOM+NoA & & 0.445 & \textbf{1.000} & 0.797 & 0.491 \\ \cline{3-8} 
                 & & \textbf{GloCOM} & & \textbf{0.453} & 0.956 & \textbf{0.806} & \textbf{0.502} \\ \hline
        \end{tabular}
        }
        \caption{Ablation study with $K = 50$ on GoogleNews and SearchSnippets datasets. Due to space limitations, we report the standard deviation in the Appendix~\ref{appendix:additional-results}.}
        \label{tab:ablation-study}
    \end{table}

    We conducted experiments to assess the efficacy of our approach compared to other baselines. Table~\ref{tab:main} presents the overall topic quality and document-topic distribution quality across four datasets: GoogleNews, SearchSnippets, StackOverflow, and Biomedical, for 50 and 100 topics. From the results, our approach demonstrates superiority over the other state-of-the-art models in terms of text clustering, which validates the effectiveness of our method. By deriving a local text document from its aggregated global clustering document, GloCOM can greatly enhance the quality of the document-topic distribution thus improving the semantic richness of the document representation.

    Furthermore, the competitive TC results of GloCOM compared to kNNTM demonstrate the benefits of our global short text aggregation in providing unbiased training for handling the label sparsity issue. This suggests that documents within a cluster have more semantically similar words compared to those using kNN optimal transport-based distance.  It is worth noting that, while the ECRTM model has a high TD in some settings, its TC score and text clustering results are notably lower than those of other short text methods. Please see Appendix~\ref{appendix:example-topics} for examples of the discovered topics.

\subsection{Ablation Study} \label{subsec:ablation}

    We analyze the effects of different variants of GloCOM, including (i) ECRTM, the base model of GloCOM; (ii) GloCOM+NoA, a GloCOM without global augmentation for input document; and (iii) GloCOM+NoC, a GloCOM without global clustering context, on both GoogleNews and SearchSnippets under 50 topics. As shown in Table~\ref{tab:ablation-study}, the improvements from the ECRTM model to other GloCOM variants demonstrate the two modules' effectiveness. Moreover, the model incorporating both modules achieves the highest performance in terms of document-topic distribution quality. Regarding topic quality, the GloCOM model obtains the best TC score while compromising the TD score when not using global augmentation.

\subsection{Clustering Effectiveness}

    \begin{figure}
        \centering
        \includegraphics[width=\linewidth]{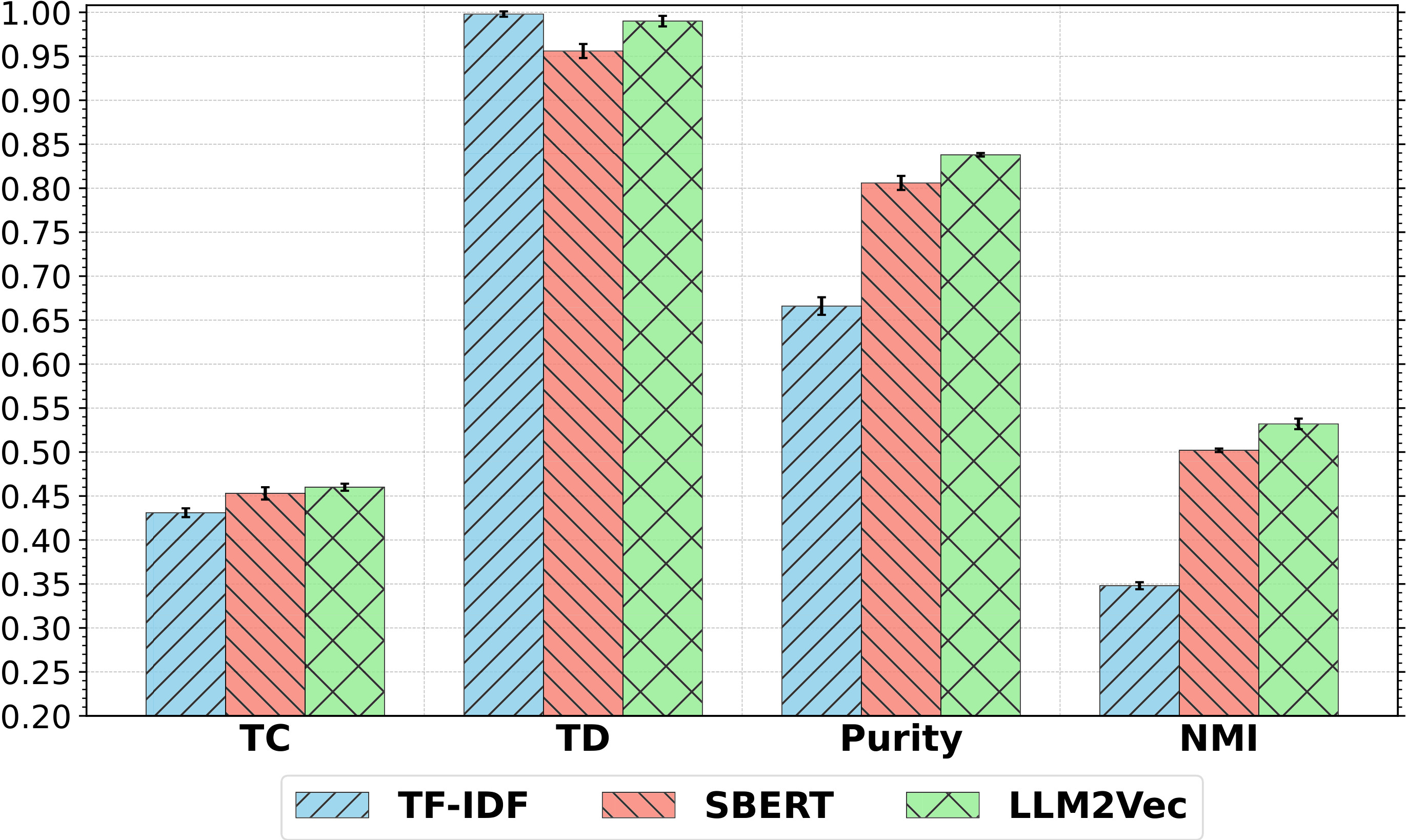}
        \vspace{-1.57em}
        \caption{Clustering effectiveness of the GloCOM model with different representations for short text clustering ($K = 50$) on the SearchSnippets dataset.}
        \label{fig:cluster_effectiveness}
    \end{figure}

    \begin{figure*}[t]
    \centering
    \resizebox{0.75\textwidth}{!}{  
        \begin{tabular}{ccc}  
            \begin{subfigure}[b]{0.33\textwidth}
                \centering
                \includegraphics[width=\textwidth]{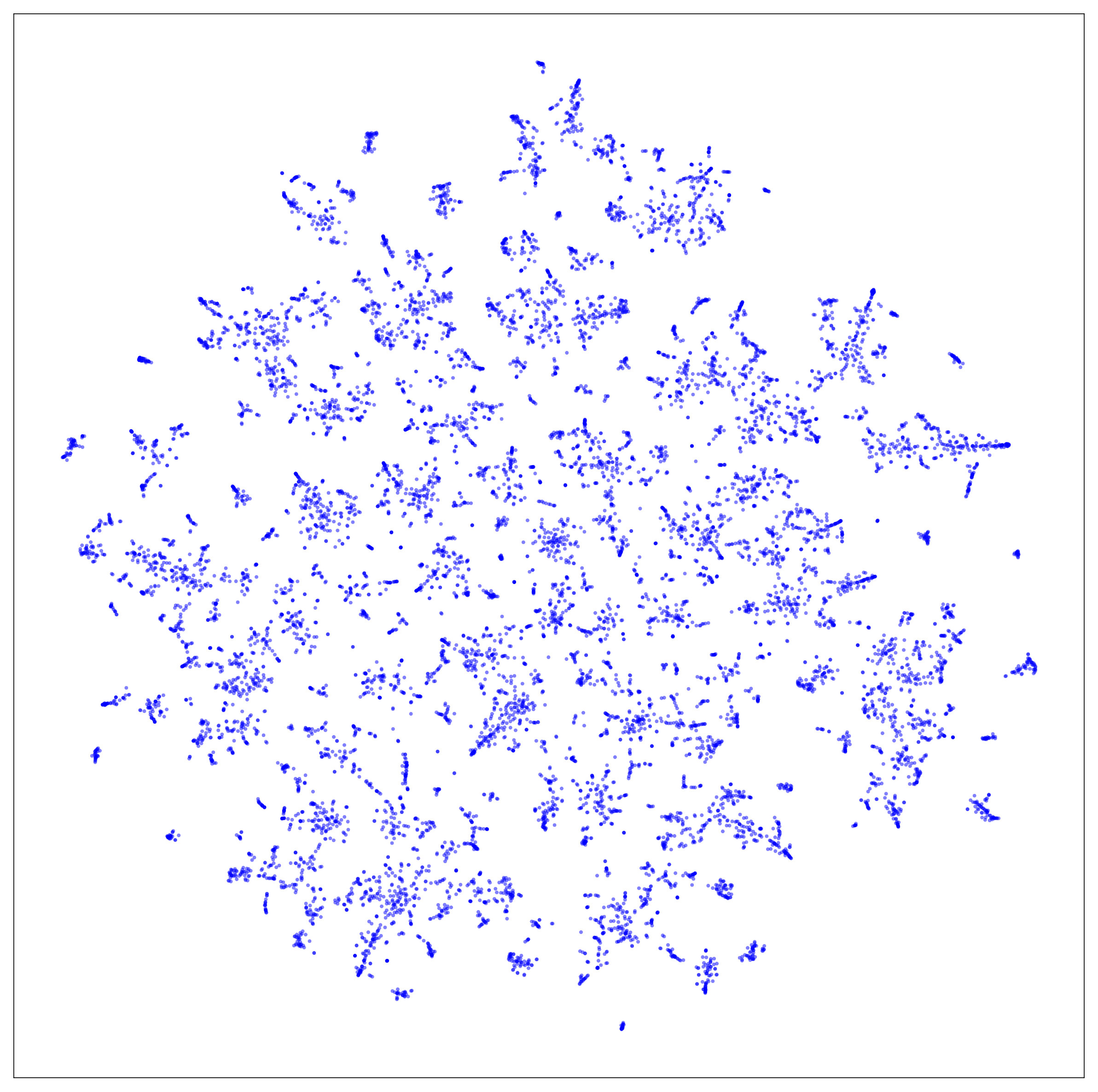}
                \caption{TSCTM}
                \label{fig:TSCTM-topic-distribution}
            \end{subfigure}
            &
            \begin{subfigure}[b]{0.33\textwidth}
                \centering
                \includegraphics[width=\textwidth]{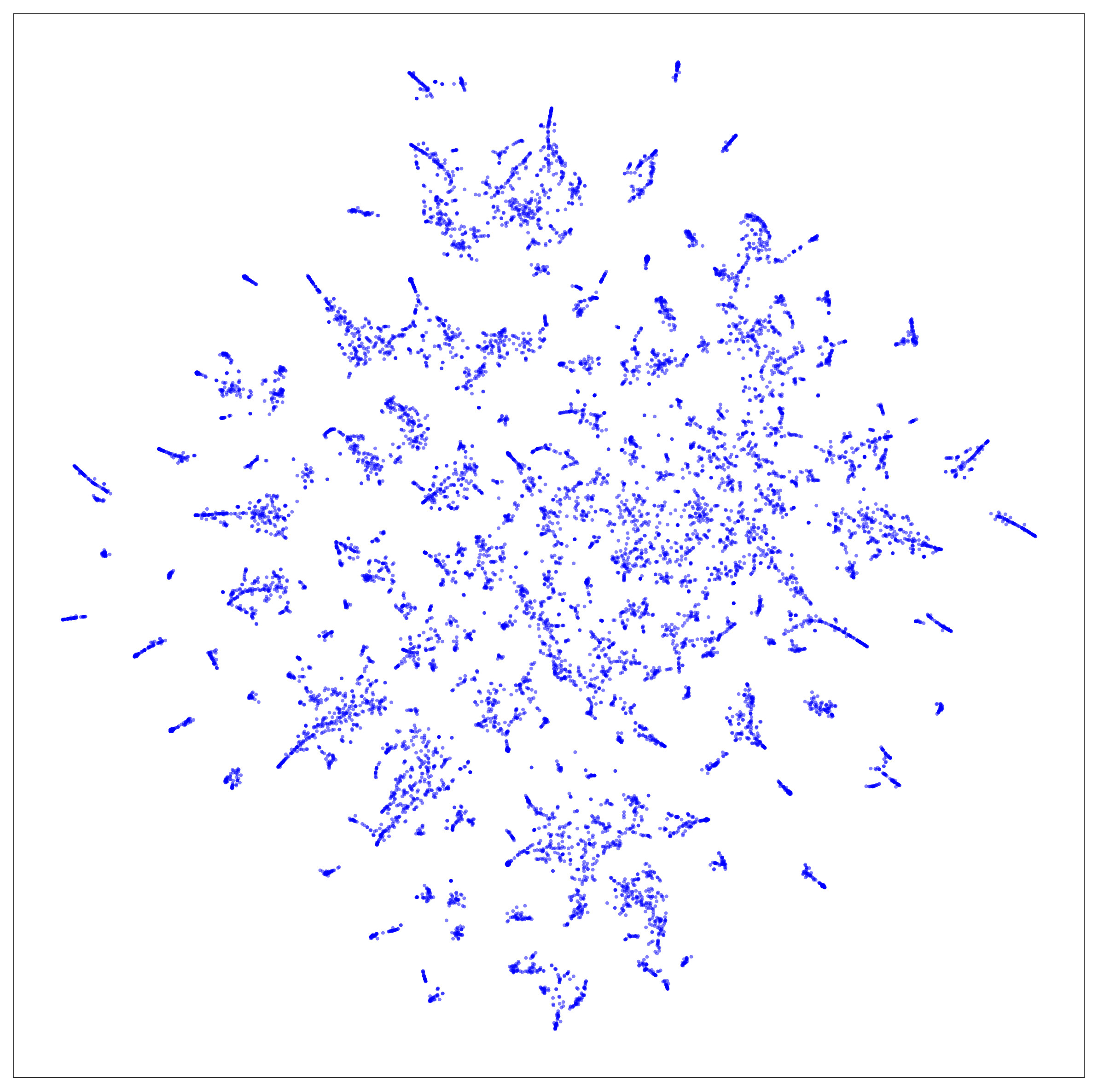}
                \caption{kNNTM}
                \label{fig:KNNTM-topic-distribution}
            \end{subfigure}
            &
            \begin{subfigure}[b]{0.33\textwidth}
                \centering
                \includegraphics[width=\textwidth]{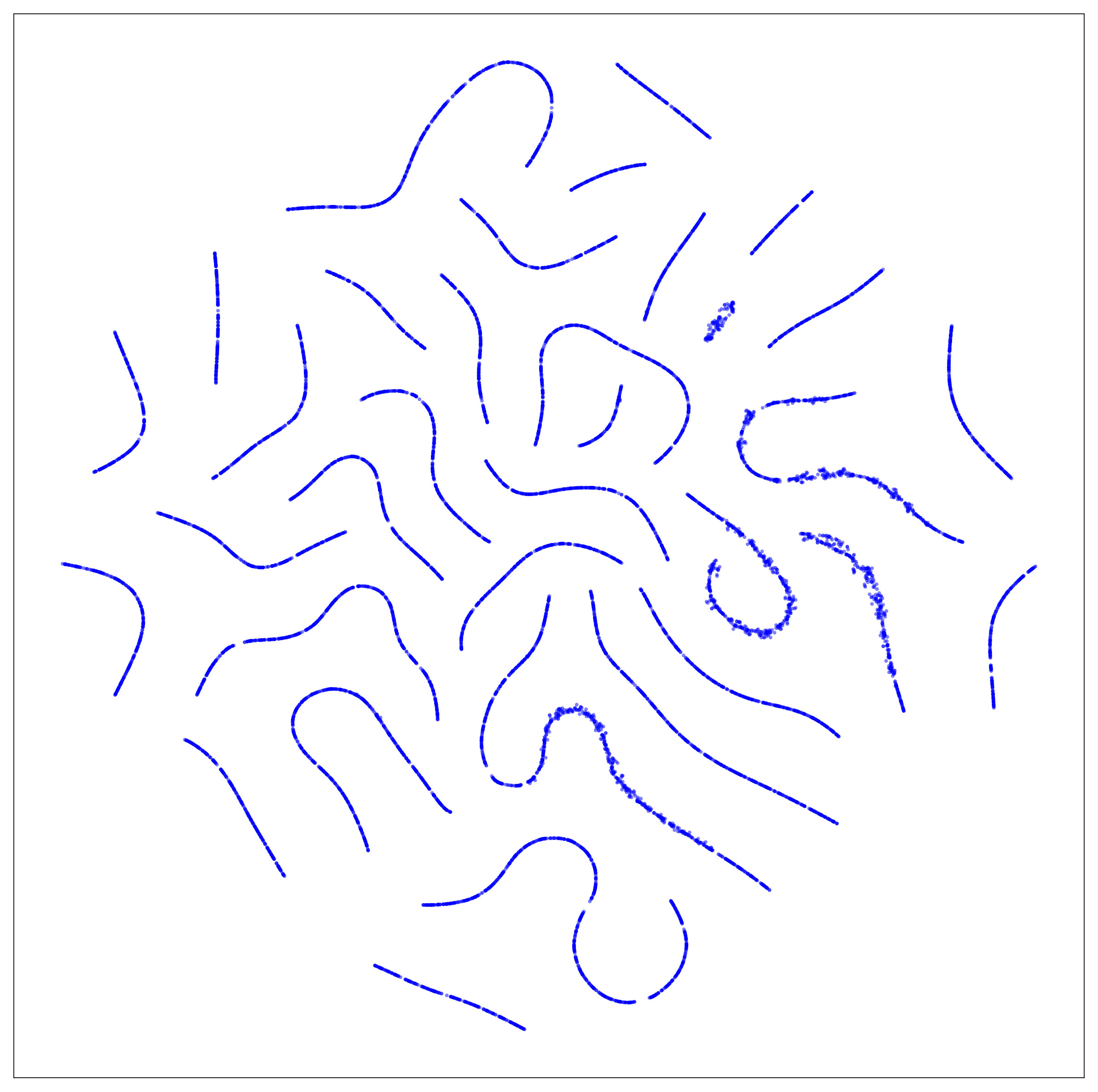}
                \caption{GloCOM}
                \label{fig:GloCOM-topic-distribution}
            \end{subfigure}
        \end{tabular}
    }
    \caption{The t-SNE visualization shows the topic distributions learned by various short text models.}
    \label{fig:topic-distribution-tsne-viz}
    \end{figure*}

    To demonstrate the necessity of pre-trained language model (PLM) embeddings for clustering in our approach, we conducted experiments on different representations including (i) \(\text{TF-IDF}\), a common representation used in traditional self-aggregation topic model~\cite{Quan_aggregate_2015}; (ii) SBERT, a PLM-based embedding representation adopted from BERT~\cite{reimers-gurevych-2019-sbert}; (iii) LLM2Vec, a PLM-based representation harnessing the power of large language models~\cite{behnamghader2024llm2veclargelanguagemodels}. As shown in Figure~\ref{fig:cluster_effectiveness}, incorporating  PLM-based embedding models notably improves the effectiveness of our method compared to the TF-IDF model. These results prove the efficacy and necessity of using PLM-based representation, as discussed in Section~\ref{subsec:short-text-aggre}.

\subsection{Topic Distribution Analysis} \label{subsec:topic-distribution-analysis}

    Figure~\ref{fig:topic-distribution-tsne-viz} displays the t-SNE~\cite{maaten_2008tsne} visualization for learned document-topic distributions with $K = 50$ on the SearchSnippets short text dataset. It is evident that compared to TSCTM and kNNTM, the samples in GloCOM are more aggregated and more distinctly spread across the space. These clear separations and divisions demonstrate the effectiveness of our method by generating local topic distributions from their clustered global distribution. This also explains GloCOM's competitive topic quality performance.

\section{Related Work} \label{sec:related-work}

    \paragraph{Standard Topic Modeling.} 
     Traditional topic models like LDA~\cite{blei2003lda} and probabilistic LSI~\cite{1999plsi} rely on generative probabilistic frameworks. Despite improvements~\cite{2006dtm, li_2015_group, NGUYEN2022bsp}, they remain inefficient and underperform compared to modern neural network-based approaches, especially those using VAE architectures~\cite{kingma2013vae}. Current advancements include integrating pre-trained language models~\cite{bianchi-etal-2021-ctm, han-etal-2023-utopic, pham2024neuromaxenhancingneuraltopic}, leveraging optimal transport distances~\cite{zhao2020nstm}, and applying contrastive loss~\cite{2021contrastiventm}. Other methods refine the generative process by using pre-trained embeddings~\cite{dieng2020etm, 2022hyperminer}, or optimal transport distances~\cite{wang2022wete, wu2023effective, wu2024fastopicfastadaptivestable}. However, these models still struggle with short texts due to data and label sparsity problems, as shown in Section~\ref{sec:exp}. While some studies use large language models to describe topics as descriptions~\cite{pham2023topicgpt}, this differs from the original LDA framework used in this paper, which infers topic-word distributions.

    \paragraph{Short Text Topic Modeling.} Conventional short text topic models~\cite{Li_gmm_2016, Li_gmm_2017, Yin_gmm_2014} assume each text is generated by a few topics, while Biterm Topic Models~\cite{biterm_shortTM, Cheng_BTM} incorporate co-occurrence patterns. Aggregation techniques~\cite{Hong_2010, Tang_shorttm_2013, Quan_aggregate_2015} have also been used to mitigate data sparsity. However, these methods have drawbacks, such as the difficulty of inferring individual document topics~\cite{Weng_twitter_2010} and high computational costs~\cite{Zuo2016TopicMO}. Clustering is also ineffective, as term frequency representations fail to capture short text semantics~\cite{Jin_cluster_2011}.  
    
    Neural short text topic models have recently outperformed traditional models in both performance and generalization~\cite{wu2024survey}. Some use pre-trained embeddings~\cite{dieng2020etm, bianchi-etal-2021-ctm} or word co-occurrence graphs~\cite{zhu-etal-2018-graphbtm, wang-etal-2021-graph-short-topics}, while others focus on variable-length corpora~\cite{NEURIPS2022_mctm}. Topic distribution quantization~\cite{wu-etal-2020-short, wu-etal-2022-mitigating} has shown to effectively handle data sparsity, with kNNTM~\cite{lin-etal-2024-combating} pioneering label sparsity solutions. Although kNNTM is state-of-the-art for short documents, its use of optimal transport distances increases computational costs compared to our global aggregation clustering approach.

\section{Conclusion} \label{sec:conclusion}

   In this paper, we propose a novel neural model for short text topic modeling, called GloCOM, which leverages aggregated global clustering context and integrates this context into the reconstruction target. Our experiments demonstrate that our method effectively addresses both data and label sparsity challenges, outperforming state-of-the-art models and producing high-quality topics and document representations for short texts.

\section*{Limitations}


    Although our approach demonstrates promising results in addressing sparsity issues in short text topic modeling, there are a few limitations to acknowledge. Firstly, the method has to determine the number of groups for creating global documents. Future research can focus on automatically selecting the optimal number of clusters. Secondly, the method's reliance on clustering with pre-trained language models makes it challenging to apply in dynamic or real-time settings. Addressing this limitation will require further investigation into adapting both clustering and topic modeling for dynamic environments.

\section*{Ethical Considerations}
   We adhere to the ACL Code of Ethics and the terms of each codebase license. Our method aims to advance the field of topic modeling, and we are confident that, when used properly and with care, it poses no significant social risks.
    
\bibliography{acl_latex}
\appendix

\section{GloCOM Algorithm} \label{appendix:algorithm} 
    \begin{algorithm}[H]
    \caption{GloCOM training procedure.}
    \begin{algorithmic}[1]
    \REQUIRE Input corpus $\mathbf{X}$, Topic number $K$, epoch number $N$, and clusters $G$.
    \ENSURE $K$~topic-word distributions $\beta_k$, $N$ doc-topic distributions $\theta_d^g$
    \FOR{epoch from 1 to $N$}
        \STATE For a random batch of $B$ documents do
        \STATE $\mathcal{L}_{\text{batch}} \leftarrow 0$;
        \FOR{each local doc $x^d$ and its respective global doc $x^g$ in the batch}
            \STATE Compute the adaptive variable $p_d$;
            \STATE Compute the global topic distribution $\theta^g$;
            \STATE Compute the local topic distribution $\theta^g_d$ by Eq.~\ref{eq:generate-topic-distribution};
            \STATE $\mathcal{L}_{\text{batch}} \leftarrow \mathcal{L}_{\text{batch}} + \mathcal{L}_{\text{GloCOM}}$ by Eq.~\ref{eq:overall};
        \ENDFOR
        \STATE Update model parameters with $\nabla \mathcal{L}_{\text{batch}}$;
    \ENDFOR
    \end{algorithmic}
    \end{algorithm}
    
\section{Embedding Clustering Regularization}\label{appendix:ecr}
   Proposed by~\cite{wu2023effective}, Embedding Clustering Regularization (ECR) ensures that each topic embedding serves as the center of its distinct word embedding cluster in the semantic space. Specifically, it leverages optimal transport (OT) distance~\cite{Peyr2018ComputationalOT} as follows: 
        \begin{equation*} \label{eq:ecr}
        \begin{split}
            &\mathcal{L}_{\mathrm{ECR}} = \sum_{i=1}^V \sum_{j=1}^K \Vert \mathbf{w}_i-\mathbf{t}_j \Vert^2  \psi^*_{ij},
        \end{split}
        \end{equation*}
    with $\psi^*$ is the solution to the following OT optimization problem:
        \begin{equation*} \label{eq:ot-ecr}
        \begin{split}
            \text{minimize} & \  \langle C_{\mathrm{WT}}, \psi \rangle - \nu H(\psi) \\
            \text{s.t.} &  \ \psi \in \mathbb{R}^{V\times K} \\
             & \ \psi \mathds{1}_K = \frac{1}{V} \mathds{1}_V, \psi^T \mathds{1}_V = \frac{1}{K} \mathds{1}_K, \\
        \end{split}
        \end{equation*}
    where $C_{\mathrm{WT}} \in \mathbb{R}^{V \times K}$ represents the distance matrix between word embeddings and topic embeddings. The optimal transport matrix $\psi^*$ is derived using the Sinkhorn algorithm. For two matrices $X, Y$ with the same size, $\langle X, Y \rangle = \sum_{i,j} X_{ij} Y_{ij}$. $H(P) = -\langle P, \log P - 1 \rangle = - \sum_{i,j} P_{ij} (\log P_{ij} - 1)$ is the Shannon entropy of $P$~\cite{2013sinkhorn}.

\section{Experiment Details}

\subsection{Dataset Statistics} \label{appendix:data-statistics}

    We evaluate the performance of our GloCOM model and other baselines using four benchmark datasets for short text topic modeling. The specifics of each dataset are as follows:
    
    \begin{itemize}[noitemsep]
        \item \textbf{GoogleNews} includes 11,109 article titles related to 152 events, originally published and processed by~\cite{yin2016model}.
        \item \textbf{SearchSnippets} consists of 12,340 snippets extracted from web searches, categorized into 8 groups by~\cite{phan2008learning}.
        \item \textbf{StackOverflow} is the dataset used in a Kaggle challenge\footnote{\url{https://www.kaggle.com/c/predict-closed-questions-on-stack-overflow}}. For this work, 20,000 question titles from 20 categories were randomly selected by~\cite{xu2017self}.
        
        \item  \textbf{Biomedical} is a subset of PubMed data provided by BioASQ\footnote{\url{http://participants-area.bioasq.org/}}, with 20,000 paper titles randomly selected from 20 categories by~\cite{xu2017self}.
    \end{itemize}

    We then aim to replicate the experimental settings established by~\cite{lin-etal-2024-combating}. We first obtain preprocessed versions of four datasets provided by the STTM library\footnote{\url{https://github.com/qiang2100/STTM}}~\cite{qiang2018STTP}. For each dataset, we then remove words with a frequency below 3 from the corpus and the vocabulary. After that, we filter out all documents with a term length of less than 2. These preprocessing steps are implemented using TopMost\footnote{\url{https://github.com/bobxwu/topmost}}.  
    
    For global clustering, these texts are embedded into a semantic representation by the common pre-trained language model, \texttt{all-MiniLM-L6-v2}\footnote{\url{https://huggingface.co/sentence-transformers/all-MiniLM-L6-v2}}. Then, these embeddings are clustered into a chosen number of groups based on the dataset characteristics using K-Means~\cite{lloyd1982least}. Table~\ref{tab:dataset-statistic} provides an overview of the dataset statistics after preprocessing.

    \begin{table}
        \centering
        \setlength{\tabcolsep}{0.7mm}
            \renewcommand{\arraystretch}{1.2}
            \resizebox{\linewidth}{!}{
        \begin{tabular}{lcccccc}
        \hline
        \textbf{Dataset} & &
          \begin{tabular}[c]{@{}c@{}}\# of\\[-0.7ex] texts\end{tabular} &
          \begin{tabular}[c]{@{}c@{}}average \\[-0.7ex] text length\end{tabular} &
          \begin{tabular}[c]{@{}c@{}}\# of \\[-0.7ex] labels\end{tabular} &
          \begin{tabular}[c]{@{}c@{}}vocab\\[-0.7ex] size\end{tabular} &
          \begin{tabular}[c]{@{}c@{}}\# of \\[-0.7ex] groups\end{tabular} \\ \hline
        GoogleNews     & & 11,019 & 5.753  & 152 & 3,473 & 200 \\
        SearchSnippets & & 12,294 & 14.426 & 8   & 4,618 & 40  \\
        StackOverflow  & & 16,378 & 4.988  & 20  & 2,226 & 40  \\
        Biomedical     & & 19,433 & 7.430  & 20  & 3,867 & 40  \\ \hline
        \end{tabular}
        }
        \caption{Dataset statistics after preprocessing.}
        \label{tab:dataset-statistic}
    \end{table}

\subsection{Model Implementation} \label{appendix:model-implementation} 
    All experiments are conducted on two GeForce RTX 3090 24GB GPUs with CUDA version 12.5, using PyTorch 2.4.1 in a Python 3.10.14 environment. For our GloCOM model, it takes less than 0.5 hours to complete the training process per setting. We utilize the same encoder network settings for both global document distribution ($\phi$) and local adapter ($\gamma$) inference networks. Following the architecture in~\cite{wu2023effective}, both networks comprise a two-layer softplus-activated MLP and an additional layer for the mean and covariance of the latent variable for a fair comparison. We set $\tau = 0.2$ in Eq.~\ref{eq:beta_decomposition} as default by~\cite{wu2023effective}. We train our model for $N = 200$ epochs with a batch size of 200, applying the common Adam optimizer~\cite{kingma2017adam} with a learning rate of 0.002. The other hyperparameters, $\eta$ — the augmentation coefficient, $\epsilon$ — the prior variance factor of the adaptive variable, and $\lambda_{\mathrm{ECR}}$ — the ECR weight hyperparameter, are selected from a range as follows:
    
    \begin{itemize}
        \item $\eta \in [0.01, 0.05, 0.1, 0.5, 1.0]$
        \item $\epsilon \in [0.001, 0.01, 0.1]$
        \item $\lambda_{\mathrm{ECR}} \in [10, 20, 30, 60, 90]$
    \end{itemize}
     \begin{table*}
        \centering
        \setlength{\tabcolsep}{1.1mm}
        \renewcommand{\arraystretch}{1.25}
        \begin{tabular}{llcccc}
        \hline
        \multicolumn{2}{c}{\textbf{Method}} & TC                   & TD                   & Purity               & NMI                  \\ \hline
        \multirow{4}{*}{\textbf{\begin{tabular}[c]{@{}l@{}}Google\\ News\end{tabular}}}     & ECRTM & 0.441±0.003 & 0.987±0.023 & 0.396±0.007 & 0.615±0.011 \\
                 & GloCOM+NoC          & 0.465±0.010          & 0.989±0.010          & 0.548±0.013          & 0.768±0.011          \\
                 & GloCOM+NoA         & 0.436±0.001          & \textbf{1.000±0.000} & 0.548±0.013          & 0.770±0.004          \\ \cline{2-6} 
                 & \textbf{GloCOM}           & \textbf{0.475±0.009} & 0.999±0.001          & \textbf{0.586±0.012} & \textbf{0.817±0.002} \\ \hline
        \multirow{4}{*}{\textbf{\begin{tabular}[c]{@{}l@{}}Search\\ Snippets\end{tabular}}} & ECRTM & 0.450±0.006 & 0.998±0.002 & 0.711±0.020 & 0.419±0.022 \\
                 & GloCOM+NoC          & 0.449±0.003          & 0.855±0.028          & 0.805±0.008          & 0.484±0.001          \\
                 & GloCOM+NoA         & 0.445±0.008          & \textbf{1.000±0.000} & 0.797±0.011          & 0.491±0.004          \\ \cline{2-6} 
                 & \textbf{GloCOM}           & \textbf{0.453±0.007} & 0.956±0.008          & \textbf{0.806±0.008} & \textbf{0.502±0.001} \\ \hline
        \end{tabular}
        \caption{Full ablation study results with $K = 50$ on GoogleNews and SearchSnippets datasets. We report the mean and the standard deviation of three different runs.}
        \label{tab:full-ablation-study}
    \end{table*}
    
\subsection{kNNTM Implementation} \label{appendix:kNNTM-implementation}
    The kNNTM~\cite{lin-etal-2024-combating} model is the state-of-the-art short text topic modeling baseline. However, the codebase is not publicly available, so we replicate the implementation of kNNTM with exact settings for the model and for each dataset as described in~\cite{lin-etal-2024-combating}. For the additional Biomedical dataset, we use the same settings as the SearchSnippets dataset.
    
   It is worth noting that calculating the optimal transport distance required for pairwise document comparisons is costly and time-consuming. It takes about two weeks on four NVIDIA A100 80GB GPUs to measure the distances for four datasets, even though we use the Sinkhorn algorithm~\cite{2013sinkhorn} with a batch implementation. We will publish the code for the kNNTM models alongside our codebase.

\section{Additional Results} \label{appendix:additional-results}
    Due to space constraints, the standard deviations of the results in the Section~\ref{sec:exp} are not included in the main paper. Here, we present the means and standard deviations of the results in Tables~\ref{tab:main-50-full},~\ref{tab:main-100-full}, and~\ref{tab:full-ablation-study}, which correspond to Tables~\ref{tab:main} and~\ref{tab:ablation-study} of the main paper.

    \begin{table*}
        \centering
        \setlength{\tabcolsep}{1mm}
        \renewcommand{\arraystretch}{1.2}
        \resizebox{\linewidth}{!}{
        \begin{tabular}{lccccccccccc}
        \hline
        \multirow{2}{*}{\textbf{Model}} & & \multicolumn{4}{c}{\textbf{GoogleNews}} & & \multicolumn{4}{c}{\textbf{SearchSnippets}} \\ \cline{3-6} \cline{8-11}
                                        & & TC          & TD          & Purity      & NMI         & & TC          & TD          & Purity      & NMI         \\ \hline
        ProdLDA                         & & 0.437±0.001 & 0.991±0.003 & 0.201±0.011 & 0.384±0.031 & & 0.406±0.007 & 0.546±0.039 & 0.731±0.017 & 0.435±0.009 \\
        ETM                             & & 0.402±0.004 & 0.916±0.006 & 0.366±0.032 & 0.560±0.030 & & 0.397±0.003 & 0.594±0.012 & 0.688±0.002 & 0.389±0.013 \\
        ECRTM                           & & 0.441±0.003 & 0.987±0.023 & 0.396±0.007 & 0.615±0.011 & & 0.450±0.006 & \textbf{0.998±0.002} & 0.711±0.020 & 0.419±0.022 \\
        FASTopic                        & & 0.446±0.010 & 0.440±0.020 & 0.351±0.006 & 0.659±0.006 & & 0.395±0.001 & 0.710±0.019 & 0.792±0.008 & 0.481±0.008 \\
        NQTM                            & & 0.408±0.003 & 0.959±0.002 & 0.536±0.005 & 0.716±0.010 & & 0.436±0.001 & 0.922±0.003 & 0.435±0.007 & 0.150±0.007 \\
        TSCTM                           & & 0.437±0.005 & 0.988±0.003 & 0.552±0.009 & 0.761±0.010 & & 0.424±0.003 & 0.993±0.007 & 0.724±0.005 & 0.386±0.006 \\
        KNNTM                           & & 0.435±0.006 & 0.986±0.006 & 0.579±0.010 & 0.795±0.007 & & 0.425±0.004 & 0.995±0.003 & 0.768±0.001 & 0.429±0.001 \\ \hline
        \textbf{GloCOM}                 & & \textbf{0.475±0.009} & \textbf{0.999±0.001} & \textbf{0.586±0.012} & \textbf{0.817±0.002} & & \textbf{0.453±0.007} & 0.956±0.008 & \textbf{0.806±0.008} & \textbf{0.502±0.001} \\ \hline
        \multirow{2}{*}{\textbf{Model}} & & \multicolumn{4}{c}{\textbf{StackOverflow}} & & \multicolumn{4}{c}{\textbf{Biomedical}} \\ \cline{3-6} \cline{8-11}
                                        & & TC          & TD          & Purity      & NMI         & & TC          & TD          & Purity      & NMI         \\ \hline
        ProdLDA                         & & 0.388±0.006 & 0.588±0.011 & 0.117±0.023 & 0.151±0.025 & & 0.469±0.009 & 0.520±0.054 & 0.136±0.014 & 0.177±0.014 \\
        ETM                             & & 0.367±0.006 & 0.766±0.015 & 0.418±0.012 & 0.280±0.005 & & 0.450±0.008 & 0.723±0.023 & 0.406±0.003 & 0.273±0.005 \\
        ECRTM                           & & 0.381±0.006 & 0.941±0.035 & 0.197±0.024 & 0.192±0.026 & & 0.468±0.005 & 0.987±0.013 & 0.414±0.005 & 0.315±0.005 \\
        FASTopic                        & & 0.317±0.010 & 0.222±0.021 & 0.408±0.007 & 0.486±0.011 & & 0.418±0.007 & 0.482±0.017 & 0.456±0.003 & 0.369±0.004 \\
        NQTM                            & & 0.382±0.002 & 0.933±0.008 & 0.392±0.016 & 0.238±0.007 & & 0.471±0.007 & 0.915±0.022 & 0.191±0.011 & 0.109±0.006 \\
        TSCTM                           & & 0.378±0.005 & 0.911±0.011 & 0.572±0.009 & 0.418±0.002 & & 0.484±0.006 & 0.972±0.009 & 0.480±0.008 & 0.341±0.004 \\
        KNNTM                           & & 0.380±0.005 & 0.922±0.009 & 0.636±0.005 & 0.490±0.005 & & \textbf{0.490±0.008} & 0.972±0.008 & 0.526±0.011 & 0.380±0.007 \\ \hline
        \textbf{GloCOM}                 & & \textbf{0.390±0.012} & \textbf{0.962±0.006} & \textbf{0.653±0.002} & \textbf{0.588±0.002} & & \textbf{0.490±0.005} & \textbf{0.998±0.003} & \textbf{0.546±0.005} & \textbf{0.437±0.004} \\ \hline
        \end{tabular}
        }
        \caption{Full topic quality results, measured using TC and TD, and document-topic distribution quality, assessed with NMI and Purity with $K = 50$. The \textbf{bold} values indicate the best performance. We report the mean and the standard deviation of three different runs.}
        \label{tab:main-50-full}
        \end{table*}
        
    \begin{table*}
        \centering
        \setlength{\tabcolsep}{1mm}
        \renewcommand{\arraystretch}{1.2}
        \resizebox{\linewidth}{!}{
        \begin{tabular}{lccccccccccc}
        \hline
        \multirow{2}{*}{\textbf{Model}} & & \multicolumn{4}{c}{\textbf{GoogleNews}} & & \multicolumn{4}{c}{\textbf{SearchSnippets}} \\ \cline{3-6} \cline{8-11}
                                        & & TC          & TD          & Purity               & NMI         & & TC          & TD          & Purity      & NMI         \\ \hline
        ProdLDA                         & & 0.435±0.007 & 0.611±0.015 & 0.611±0.015          & 0.600±0.046 & & 0.424±0.005 & 0.679±0.018 & 0.766±0.004 & 0.415±0.004 \\
        ETM                             & & 0.398±0.001 & 0.677±0.018 & 0.554±0.014          & 0.713±0.012 & & 0.389±0.001 & 0.448±0.012 & 0.692±0.012 & 0.365±0.013 \\
        ECRTM                          & & 0.418±0.004 & \textbf{0.991±0.006} & 0.342±0.012 & 0.491±0.013 & & 0.432±0.003 & \textbf{0.966±0.006} & 0.789±0.006 & 0.443±0.002 \\
        FASTopic                       & & 0.438±0.011 & 0.369±0.025 & 0.458±0.010          & 0.722±0.010 & & 0.386±0.008 & 0.634±0.017 & 0.807±0.015 & 0.458±0.010 \\
        NQTM                           & & 0.397±0.001 & 0.898±0.010 & 0.706±0.003          & 0.788±0.001 & & 0.438±0.004 & 0.638±0.006 & 0.334±0.011 & 0.077±0.005 \\
        TSCTM                          & & 0.448±0.002 & 0.941±0.008 & 0.754±0.001          & 0.835±0.002 & & 0.430±0.005 & 0.894±0.020 & 0.757±0.008 & 0.384±0.004 \\
        KNNTM                          & & 0.441±0.003 & 0.959±0.004 & \textbf{0.797±0.010} & 0.870±0.001 & & 0.421±0.001 & 0.948±0.006 & 0.800±0.006 & 0.421±0.002 \\ \hline
        \textbf{GloCOM}                 & & \textbf{0.450±0.006} & 0.944±0.007 & 0.761±0.012 & \textbf{0.900±0.003} & & \textbf{0.443±0.001} & 0.920±0.002 & \textbf{0.822±0.001} & \textbf{0.501±0.003} \\ 
        \hline
        \multirow{2}{*}{\textbf{Model}} & & \multicolumn{4}{c}{\textbf{StackOverflow}} & & \multicolumn{4}{c}{\textbf{Biomedical}} \\ \cline{3-6} \cline{8-11}
                                        & & TC                   & TD                   & Purity      & NMI         & & TC                   & TD          & Purity      & NMI         \\ \hline
        ProdLDA                         & & \textbf{0.382±0.001} & 0.466±0.019          & 0.098±0.001 & 0.090±0.015 & & 0.463±0.005          & 0.465±0.040 & 0.079±0.005 & 0.050±0.006 \\
        ETM                             & & 0.369±0.001          & 0.444±0.006          & 0.475±0.020 & 0.331±0.017 & & 0.452±0.004          & 0.476±0.009 & 0.404±0.007 & 0.268±0.006 \\
        ECRTM                           & & 0.375±0.003          & \textbf{0.993±0.004} & 0.172±0.013 & 0.179±0.016 & & 0.444±0.001          & 0.974±0.007 & 0.124±0.007 & 0.113±0.008 \\
        FASTopic                        & & 0.309±0.005          & 0.186±0.012          & 0.495±0.013 & 0.514±0.016 & & 0.440±0.002          & 0.457±0.040 & 0.495±0.013 & 0.375±0.011 \\
        NQTM                            & & 0.379±0.001          & 0.818±0.006          & 0.417±0.003 & 0.255±0.002 & & 0.460±0.007          & 0.572±0.046 & 0.142±0.002 & 0.056±0.002 \\
        TSCTM                           & & 0.380±0.004          & 0.620±0.010          & 0.563±0.007 & 0.386±0.006 & & \textbf{0.485±0.005} & 0.806±0.009 & 0.487±0.004 & 0.330±0.003 \\
        KNNTM                           & & 0.381±0.003          & 0.663±0.016          & 0.611±0.013 & 0.436±0.003 & & 0.483±0.001         & 0.848±0.006 & 0.530±0.003 & 0.362±0.001 \\ \hline
        \textbf{GloCOM}                 & & \textbf{0.382±0.007} & 0.804±0.004 & \textbf{0.658±0.002} & \textbf{0.585±0.002} & & 0.462±0.008 & \textbf{0.997±0.002}  & \textbf{0.536±0.007} & \textbf{0.422±0.005} \\ \hline
        \end{tabular}
        }
        \caption{Full topic quality results, measured using TC and TD, and document-topic distribution quality, assessed with NMI and Purity with $K = 100$. The \textbf{bold} values indicate the best performance. We report the mean and the standard deviation of three different runs.}
        \label{tab:main-100-full}
    \end{table*}

\section{Examples of Discovered Topics} \label{appendix:example-topics}
    We provide the first 20 discovered topics of our GloCOM models from the SearchSnippets dataset under $K = 50$. As shown in Table~\ref{tab:discovered-topics}, the model can identify meaningful topics, such as Topic \#2, which is about ideology, with relevant words like ``capitalism'', ``marxists'', and ``socialism''. Note that although the word ``wee'' appears in both Topic \#1, about sports, and Topic \#7, about entertainment, this is correct as WEE refers to World Wrestling Entertainment. This further validates our model's effectiveness in identifying high-quality topics in short text datasets.

   \begin{table*}
    \centering
    \resizebox{\linewidth}{!}{
    \begin{tabular}{l}
    \hline
    \textbf{Discovered Topic Examples} \\ \hline
    \begin{tabular}[c]{@{}l@{}}\#1: espn ncaasports mlb standings lacrosse devils playoff hockey suns tna \underline{\textbf{wwe}} afl rumors \\ 
    scores basketball \\ \#2: ideology labour capitalism marxists party socialism electoral kazakhstan radical  worker \\ taiwan democratic gates peaceful marxist \\ \#3: navy nuclear dod army mil weapons treaty nationalsecurity rand alamos corps bomb \\ atlantic naval
    invasion \\ \#4: messaging mcafee wi isps paypal voip lan measuring microprocessors supercomputing \\ connections pakistan websearch wideless scandal \\ \#5: medicare parliament bills legislation leg legislative enacted appropriations fiscal \\ representatives senate legislature opsi noaa ngdc \\ \#6: lyrics blues orchestra symphony rock pianos thurston piano bluegrass concerts midi \\ orchestras 
    bands nasoalmo solo \\ \#7: oprah entourage mainetoday comedy vhs comedies myspace askmen roberts ellen \\ metacafe \underline{\textbf{wwe}} julia rank lycos \\ \#8: messenger dom isp python kdd xmldocument verizon webbrowser markup speedtest \\ linksys fi symantec verisign safari \\ \#9: exchanges import currencies trading futures leads inflation forex commodity commodities \\ export traders exporters dollar boats \\ \#10: presenter bodybuilding resorts circuit thoroughbred ski antique forensics routines moments \\ 
    resort aspen guild democracies simpsons \\ \#11: hepatitis smoking epidemiology infections infectious prevention liver cdc cigarettes lung aids \\ 
    unaids diseases cancers cancer \\ \#12: snowboarding skiing snowboard miniclip softball miyazato sania forehand mirza candystand \\ 
    addictinggames funbrain tennis tournaments volleyball \\ \#13: aging genome biotech biomedical plants molecules majors evolutionary molecular genetics \\ 
    physiology informatics plant neuroscience biotechnology \\ \#14: astronomy physicsweb astrophysics sciam geophysics nida missions iop inventions nasa \\ 
    warming weisstein gsfc popsci physics \\ \#15: shareholder realestate nasdaq timeshare investments debt hotjobs consolidation shareholders \\ 
    bankruptcy moneycentral financials mortgage venture securities \\ \#16: geographies ivillage ucf iop athletes mens garros psychoanalysis econ professors \\ 
    pyramids arl advisors globalisation lecturers \\ \#17: identities mathematician axioms newton mathematica proofs mizar neil mathforum \\ gravitation maa  axiom solids equations isaac \\ \#18: cert speakeasy ppl maths civilizations evansville mls chakvetadze amherst portraits buenos \\ 
    sporting athens arch balls \\ \#19: merit techweb sorensen podtech cancertopics pcguide screensavers foodborne parascope \\ 
    aidsinfo coupons popsci cores optimization unveils \\ \#20: admissions doctorate gslis scholarships degree hunter grad majors graduate colleges \\ 
    degrees simmons doctoral finaid graduation\end{tabular} \\ \hline
    \end{tabular}}
    \caption{Top 15 related words of 20 discovered topics from SearchSnippets. Repeated words are \textbf{bold} and~\underline{underlined}.}
    \label{tab:discovered-topics}
    \end{table*}
\end{document}